\documentclass[letterpaper]{article}
\usepackage[preprint]{aaai2027}
\usepackage[hyphens]{url} %
\usepackage{graphicx} %
\usepackage{natbib} %
\usepackage{caption} %
\usepackage{amsmath,amssymb,booktabs,array,listings}
\usepackage{tikz}
\usetikzlibrary{arrows.meta,bending,positioning,fit,backgrounds,calc,shadows,shapes.geometric,3d,shadows.blur,fadings,shadings}
\definecolor{finblue}{HTML}{2A78D6}%
\definecolor{finorange}{HTML}{EB6834}%
\definecolor{finaqua}{HTML}{1BAF7A}%
\definecolor{finmagenta}{HTML}{C2408B}%
\definecolor{figink}{HTML}{1F2335}%
\definecolor{fingray}{HTML}{5A6374}%
\definecolor{figmid}{HTML}{6A7385}%
\definecolor{figline}{HTML}{C9D0DA}%
\definecolor{figfaint}{HTML}{EEF1F5}%
\definecolor{figok}{HTML}{1E9E6A}%
\definecolor{figbad}{HTML}{D2483E}%
\colorlet{gink}{figink}\colorlet{gmid}{figmid}\colorlet{gline}{figline}%
\colorlet{gok}{figok}\colorlet{gbad}{figbad}%
\colorlet{kink}{figink}\colorlet{kuse}{figmid}\colorlet{koff}{figline}%
\providecommand{\figTitleFont}{\sffamily\bfseries\fontsize{7.5}{9}\selectfont}%
\providecommand{\figLabelFont}{\sffamily\fontsize{7.5}{9}\selectfont}%
\providecommand{\figSmallFont}{\sffamily\fontsize{7}{8.4}\selectfont}%
\providecommand{\figBrandFont}{\sffamily\bfseries\fontsize{8.4}{10}\selectfont}%
\providecommand{\figCodeFont}{\ttfamily\fontsize{7}{8.6}\selectfont}%
\tikzset{%
  lucide/.style={line width=0.7pt, line cap=round, line join=round},%
  lu bot/.pic={\draw[lucide, pic actions, fill=none] (0.0000,0.0500) -- (0.0000,0.1000) -- (-0.0500,0.1000); \draw[lucide, pic actions, fill=none] [rounded corners=0.0250cm] (-0.1000,0.0500) rectangle (0.1000,-0.1000); \draw[lucide, pic actions, fill=none] (-0.1250,-0.0250) -- (-0.1000,-0.0250); \draw[lucide, pic actions, fill=none] (0.1000,-0.0250) -- (0.1250,-0.0250); \draw[lucide, pic actions, fill=none] (0.0375,-0.0125) -- (0.0375,-0.0375); \draw[lucide, pic actions, fill=none] (-0.0375,-0.0125) -- (-0.0375,-0.0375);},%
  lu lock/.pic={\draw[lucide, pic actions, fill=none] [rounded corners=0.0250cm] (-0.1125,0.0125) rectangle (0.1125,-0.1250); \draw[lucide, pic actions, fill=none] (-0.0625,0.0125) -- (-0.0625,0.0625) .. controls (-0.0625,0.0968) and (-0.0343,0.1250) .. (-0.0000,0.1250) .. controls (0.0343,0.1250) and (0.0625,0.0968) .. (0.0625,0.0625) -- (0.0625,0.0125);},%
  lu check/.pic={\draw[lucide, pic actions, fill=none] (0.1000,0.0750) -- (-0.0375,-0.0625) -- (-0.1000,0.0000);},%
  lu x/.pic={\draw[lucide, pic actions, fill=none] (0.0750,0.0750) -- (-0.0750,-0.0750); \draw[lucide, pic actions, fill=none] (-0.0750,0.0750) -- (0.0750,-0.0750);},%
  lu code/.pic={\draw[lucide, pic actions, fill=none] (0.0500,-0.0750) -- (0.1250,0.0000) -- (0.0500,0.0750); \draw[lucide, pic actions, fill=none] (-0.0500,0.0750) -- (-0.1250,0.0000) -- (-0.0500,-0.0750);},%
  lu file-check/.pic={\draw[lucide, pic actions, fill=none] (-0.0750,-0.1250) .. controls (-0.0887,-0.1250) and (-0.1000,-0.1137) .. (-0.1000,-0.1000) -- (-0.1000,0.1000) .. controls (-0.1000,0.1137) and (-0.0887,0.1250) .. (-0.0750,0.1250) -- (0.0250,0.1250) .. controls (0.0330,0.1250) and (0.0407,0.1218) .. (0.0463,0.1162) -- (0.0912,0.0713) .. controls (0.0968,0.0657) and (0.1000,0.0580) .. (0.1000,0.0500) -- (0.1000,-0.1000) .. controls (0.1000,-0.1137) and (0.0887,-0.1250) .. (0.0750,-0.1250) -- (-0.0750,-0.1250) -- cycle; \draw[lucide, pic actions, fill=none] (0.0250,0.1250) -- (0.0250,0.0625) .. controls (0.0250,0.0556) and (0.0306,0.0500) .. (0.0375,0.0500) -- (0.1000,0.0500); \draw[lucide, pic actions, fill=none] (-0.0375,-0.0375) -- (-0.0125,-0.0625) -- (0.0375,-0.0125);},%
  lu shield-check/.pic={\draw[lucide, pic actions, fill=none] (0.1000,-0.0125) .. controls (0.1000,-0.0750) and (0.0562,-0.1062) .. (0.0042,-0.1244) .. controls (0.0015,-0.1253) and (-0.0014,-0.1253) .. (-0.0041,-0.1242) .. controls (-0.0562,-0.1062) and (-0.1000,-0.0750) .. (-0.1000,-0.0125) -- (-0.1000,0.0750) .. controls (-0.1000,0.0819) and (-0.0944,0.0875) .. (-0.0875,0.0875) .. controls (-0.0625,0.0875) and (-0.0312,0.1025) .. (-0.0095,0.1215) .. controls (-0.0041,0.1262) and (0.0041,0.1262) .. (0.0095,0.1215) .. controls (0.0314,0.1024) and (0.0625,0.0875) .. (0.0875,0.0875) .. controls (0.0944,0.0875) and (0.1000,0.0819) .. (0.1000,0.0750) -- (0.1000,-0.0125) -- cycle; \draw[lucide, pic actions, fill=none] (-0.0375,0.0000) -- (-0.0125,-0.0250) -- (0.0375,0.0250);},%
  lu refresh-cw/.pic={\draw[lucide, pic actions, fill=none] (-0.1125,0.0000) .. controls (-0.1125,0.0617) and (-0.0617,0.1125) .. (0.0000,0.1125) .. controls (0.0314,0.1124) and (0.0616,0.1001) .. (0.0843,0.0782) -- (0.1125,0.0500); \draw[lucide, pic actions, fill=none] (0.1125,0.1125) -- (0.1125,0.0500) -- (0.0500,0.0500); \draw[lucide, pic actions, fill=none] (0.1125,0.0000) .. controls (0.1125,-0.0617) and (0.0617,-0.1125) .. (0.0000,-0.1125) .. controls (-0.0314,-0.1124) and (-0.0616,-0.1001) .. (-0.0842,-0.0782) -- (-0.1125,-0.0500); \draw[lucide, pic actions, fill=none] (-0.0500,-0.0500) -- (-0.1125,-0.0500) -- (-0.1125,-0.1125);},%
  lu sigma/.pic={\draw[lucide, pic actions, fill=none] (0.0750,0.0625) -- (0.0750,0.0875) .. controls (0.0750,0.0944) and (0.0694,0.1000) .. (0.0625,0.1000) -- (-0.0687,0.1000) .. controls (-0.0711,0.1000) and (-0.0733,0.0987) .. (-0.0743,0.0965) .. controls (-0.0754,0.0944) and (-0.0752,0.0919) .. (-0.0737,0.0900) -- (-0.0175,0.0150) .. controls (-0.0109,0.0061) and (-0.0109,-0.0061) .. (-0.0175,-0.0150) -- (-0.0737,-0.0900) .. controls (-0.0752,-0.0919) and (-0.0754,-0.0944) .. (-0.0743,-0.0965) .. controls (-0.0733,-0.0987) and (-0.0711,-0.1000) .. (-0.0687,-0.1000) -- (0.0625,-0.1000) .. controls (0.0694,-0.1000) and (0.0750,-0.0944) .. (0.0750,-0.0875) -- (0.0750,-0.0625);},%
  lu database/.pic={\draw[lucide, pic actions, fill=none] (0.0000,0.0875) ellipse (0.1125 and 0.0375); \draw[lucide, pic actions, fill=none] (-0.1125,0.0875) -- (-0.1125,-0.0875) .. controls (-0.1125,-0.1081) and (-0.0617,-0.1250) .. (0.0000,-0.1250) .. controls (0.0617,-0.1250) and (0.1125,-0.1081) .. (0.1125,-0.0875) -- (0.1125,0.0875); \draw[lucide, pic actions, fill=none] (-0.1125,0.0000) .. controls (-0.1125,-0.0206) and (-0.0617,-0.0375) .. (0.0000,-0.0375) .. controls (0.0617,-0.0375) and (0.1125,-0.0206) .. (0.1125,0.0000);},%
  lu gauge/.pic={\draw[lucide, pic actions, fill=none] (0.0000,-0.0250) -- (0.0500,0.0250); \draw[lucide, pic actions, fill=none] (-0.1082,-0.0875) .. controls (-0.1385,-0.0352) and (-0.1266,0.0319) .. (-0.0804,0.0708) .. controls (-0.0341,0.1096) and (0.0341,0.1096) .. (0.0804,0.0708) .. controls (0.1266,0.0319) and (0.1385,-0.0352) .. (0.1082,-0.0875);},%
  lu chart-column/.pic={\draw[lucide, pic actions, fill=none] (-0.1125,0.1125) -- (-0.1125,-0.0875) .. controls (-0.1125,-0.1012) and (-0.1012,-0.1125) .. (-0.0875,-0.1125) -- (0.1125,-0.1125); \draw[lucide, pic actions, fill=none] (0.0750,-0.0625) -- (0.0750,0.0375); \draw[lucide, pic actions, fill=none] (0.0125,-0.0625) -- (0.0125,0.0875); \draw[lucide, pic actions, fill=none] (-0.0500,-0.0625) -- (-0.0500,-0.0250);},%
  lu plug/.pic={\draw[lucide, pic actions, fill=none] (0.0000,-0.1250) -- (0.0000,-0.0625); \draw[lucide, pic actions, fill=none] (0.0375,0.0500) -- (0.0375,0.1250); \draw[lucide, pic actions, fill=none] (0.0625,0.0500) .. controls (0.0694,0.0500) and (0.0750,0.0444) .. (0.0750,0.0375) -- (0.0750,-0.0125) .. controls (0.0750,-0.0399) and (0.0524,-0.0625) .. (0.0250,-0.0625) -- (-0.0250,-0.0625) .. controls (-0.0524,-0.0625) and (-0.0750,-0.0399) .. (-0.0750,-0.0125) -- (-0.0750,0.0375) .. controls (-0.0750,0.0444) and (-0.0694,0.0500) .. (-0.0625,0.0500) -- (0.0625,0.0500) -- cycle; \draw[lucide, pic actions, fill=none] (-0.0375,0.0500) -- (-0.0375,0.1250);},%
  lu square-terminal/.pic={\draw[lucide, pic actions, fill=none] (-0.0625,0.0125) -- (-0.0375,0.0375) -- (-0.0625,0.0625); \draw[lucide, pic actions, fill=none] (-0.0125,-0.0125) -- (0.0375,-0.0125); \draw[lucide, pic actions, fill=none] [rounded corners=0.0250cm] (-0.1125,0.1125) rectangle (0.1125,-0.1125);},%
  lu braces/.pic={\draw[lucide, pic actions, fill=none] (-0.0500,0.1125) -- (-0.0625,0.1125) .. controls (-0.0762,0.1125) and (-0.0875,0.1012) .. (-0.0875,0.0875) -- (-0.0875,0.0250) .. controls (-0.0875,0.0113) and (-0.0988,0.0000) .. (-0.1125,0.0000) .. controls (-0.0988,0.0000) and (-0.0875,-0.0113) .. (-0.0875,-0.0250) -- (-0.0875,-0.0875) .. controls (-0.0875,-0.1013) and (-0.0762,-0.1125) .. (-0.0625,-0.1125) -- (-0.0500,-0.1125); \draw[lucide, pic actions, fill=none] (0.0500,-0.1125) -- (0.0625,-0.1125) .. controls (0.0762,-0.1125) and (0.0875,-0.1012) .. (0.0875,-0.0875) -- (0.0875,-0.0250) .. controls (0.0875,-0.0113) and (0.0987,0.0000) .. (0.1125,0.0000) .. controls (0.0988,0.0000) and (0.0875,0.0113) .. (0.0875,0.0250) -- (0.0875,0.0875) .. controls (0.0875,0.1012) and (0.0762,0.1125) .. (0.0625,0.1125) -- (0.0500,0.1125);},%
  lu book-open/.pic={\draw[lucide, pic actions, fill=none] (0.0000,0.0875) -- (0.0000,-0.1125); \draw[lucide, pic actions, fill=none] (0.1000,-0.0875) .. controls (0.1137,-0.0875) and (0.1250,-0.0762) .. (0.1250,-0.0625) -- (0.1250,0.0875) .. controls (0.1250,0.1012) and (0.1137,0.1125) .. (0.1000,0.1125) -- (0.0500,0.1125) .. controls (0.0303,0.1125) and (0.0118,0.1032) .. (0.0000,0.0875) .. controls (-0.0118,0.1032) and (-0.0303,0.1125) .. (-0.0500,0.1125) -- (-0.1000,0.1125) .. controls (-0.1137,0.1125) and (-0.1250,0.1012) .. (-0.1250,0.0875) -- (-0.1250,-0.0625) .. controls (-0.1250,-0.0762) and (-0.1137,-0.0875) .. (-0.1000,-0.0875) -- (-0.0500,-0.0875) .. controls (-0.0303,-0.0875) and (-0.0118,-0.0968) .. (0.0000,-0.1125) .. controls (0.0118,-0.0968) and (0.0303,-0.0875) .. (0.0500,-0.0875) -- (0.1000,-0.0875) -- cycle;},%
  lu file-text/.pic={\draw[lucide, pic actions, fill=none] (-0.0750,-0.1250) .. controls (-0.0887,-0.1250) and (-0.1000,-0.1137) .. (-0.1000,-0.1000) -- (-0.1000,0.1000) .. controls (-0.1000,0.1137) and (-0.0887,0.1250) .. (-0.0750,0.1250) -- (0.0250,0.1250) .. controls (0.0330,0.1250) and (0.0407,0.1218) .. (0.0463,0.1162) -- (0.0912,0.0713) .. controls (0.0968,0.0657) and (0.1000,0.0580) .. (0.1000,0.0500) -- (0.1000,-0.1000) .. controls (0.1000,-0.1137) and (0.0887,-0.1250) .. (0.0750,-0.1250) -- (-0.0750,-0.1250) -- cycle; \draw[lucide, pic actions, fill=none] (0.0250,0.1250) -- (0.0250,0.0625) .. controls (0.0250,0.0556) and (0.0306,0.0500) .. (0.0375,0.0500) -- (0.1000,0.0500); \draw[lucide, pic actions, fill=none] (-0.0250,0.0375) -- (-0.0500,0.0375); \draw[lucide, pic actions, fill=none] (0.0500,-0.0125) -- (-0.0500,-0.0125); \draw[lucide, pic actions, fill=none] (0.0500,-0.0625) -- (-0.0500,-0.0625);},%
  lu chart-line/.pic={\draw[lucide, pic actions, fill=none] (-0.1125,0.1125) -- (-0.1125,-0.0875) .. controls (-0.1125,-0.1012) and (-0.1012,-0.1125) .. (-0.0875,-0.1125) -- (0.1125,-0.1125); \draw[lucide, pic actions, fill=none] (0.0875,0.0375) -- (0.0250,-0.0250) -- (-0.0250,0.0250) -- (-0.0625,-0.0125);},%
  lu user/.pic={\draw[lucide, pic actions, fill=none] (0.0875,-0.1125) -- (0.0875,-0.0875) .. controls (0.0875,-0.0601) and (0.0649,-0.0375) .. (0.0375,-0.0375) -- (-0.0375,-0.0375) .. controls (-0.0649,-0.0375) and (-0.0875,-0.0601) .. (-0.0875,-0.0875) -- (-0.0875,-0.1125); \draw[lucide, pic actions, fill=none] (0.0000,0.0625) circle (0.0500);},%
}%
}%
\tikzset{%
  fig title/.style={font=\figTitleFont, text=figink, inner sep=0},%
  fig label/.style={font=\figLabelFont, text=figink, inner sep=0, align=center},%
  fig small/.style={font=\figSmallFont, text=figink!85, inner sep=0, align=center},%
  fig note/.style={font=\figSmallFont, text=figmid, inner sep=0, align=center},%
  fig code/.style={font=\figCodeFont, text=figink, inner sep=0},%
  fig outline/.style={line width=0.5pt},%
  fig shadow/.style={blur shadow={shadow blur steps=6, shadow xshift=0pt,%
    shadow yshift=-0.8pt, shadow opacity=16, shadow blur radius=1.8pt}},%
  fig item/.style={rounded corners=1pt},%
  fig card/.style={fill=white, draw=figline, line width=0.5pt, rounded corners=2.5pt, fig shadow},%
  fig box/.style={fill=white, draw=figline, line width=0.5pt, rounded corners=4pt, fig shadow},%
  fig arrow/.style={-{Latex[length=1.6mm,width=1.4mm]}, line width=0.8pt},%
  fig flow/.style={fig arrow, draw=figmid!70},%
  fig pass/.style={fig arrow, draw=figok},%
  fig fail/.style={fig arrow, draw=figbad!80, dash pattern=on 3pt off 1.8pt},%
  fig back/.style={fig arrow, draw=finaqua!85!black},%
  gword/.style={fig label},%
  gnote/.style={fig note},%
  gflow/.style={fig flow},%
  gshade/.style={fig shadow},%
}%
\newcommand{\NSkills}{131}

\newcommand{\NAlgorithms}{48}
\newcommand{\NAlgorithmTasks}{11}

\newcommand{\NGuards}{39}

\newcommand{\NTools}{67}
\newcommand{\NCards}{6}

\newcommand{\NMethods}{7}

\newcommand{\NTasks}{34}
\newcommand{\NTaskAlgorithms}{17}
\newcommand{\NDevTasks}{18}

\newcommand{\NHeldTasks}{16}

\newcommand{\NFamilies}{5}

\newcommand{\NRows}{128}

\newcommand{\NForcedDev}{5}
\newcommand{\NDrawn}{12}
\newcommand{\NDrawnDev}{4}

\newcommand{\NRealRows}{128}

\newcommand{\RealFirst}{31 March 2022}
\newcommand{\RealLast}{3 October 2022}
\newcommand{\RefGapMinVariance}{1.4\times10^{-6}}
\newcommand{\RefGapOther}{2.8\times10^{-14}}
\newcommand{\BudgetAttempts}{3}

\newcommand{\EvolveGenerations}{3}
\newcommand{\EvolveMinibatch}{4}

\newcommand{\StudySeeds}{3}
\newcommand{\StudyModels}{20}
\newcommand{\StudyCoreModels}{5}
\newcommand{\StudyFamilies}{8}

\newcommand{\StudyOutputCap}{2,048}

\newcommand{\FixInterfaceAlgorithms}{12}
\newcommand{\FixOutputCap}{4,096}
\newcommand{\FixEpisodesPerArm}{18}
\newcommand{\FixVisibleChecks}{6}

\newcommand{\FixAttempts}{3}
\newcommand{\FixCalls}{6}
\newcommand{\FixTokens}{12,288}
\newcommand{\FixModels}{20}
\newcommand{\FixArm}{360}

\newcommand{\FixFZeroPct}{9.7}

\newcommand{\FixFOnePct}{16.1}

\newcommand{\FixFTwoPct}{29.7}

\newcommand{\FixFThreePct}{16.1}

\newcommand{\FixFFourValid}{105}
\newcommand{\FixFFourPct}{29.2}
\newcommand{\FixFFourFirst}{61}

\newcommand{\FixFOneMinusFZeroBetter}{10}
\newcommand{\FixFOneMinusFZeroWorse}{2}

\newcommand{\FixFTwoMinusFZeroBetter}{11}

\newcommand{\FixFTwoMinusFOneBetter}{9}
\newcommand{\FixFTwoMinusFOneWorse}{4}

\newcommand{\FixFFourMinusFZeroBetter}{11}

\newcommand{\FixFOnePrograms}{962}
\newcommand{\FixFOneUsesImportsFinSkills}{892}

\newcommand{\FixFOneUsesPositions}{92}

\newcommand{\FixFFourPrograms}{881}
\newcommand{\FixFFourUsesImportsFinSkills}{859}

\newcommand{\FixFFourLastUsesPositions}{160}
\newcommand{\FixFFourLastUsesPositionsValid}{86}

\newcommand{\FixFFourLastUsesSignalPositions}{147}
\newcommand{\FixFFourLastUsesSignalPositionsValid}{16}

\newcommand{\FixSizeCut}{14B}
\newcommand{\FixLargeModels}{8}

\newcommand{\FixLargeFZeroPct}{23.6}

\newcommand{\FixLargeFTwoPct}{63.9}

\newcommand{\FixLargeFFourPct}{64.6}

\newcommand{\FixSmallModels}{12}

\newcommand{\FixRouteRight}{274}

\newcommand{\FixSmallFFourTold}{130}

\newcommand{\FixFZeroFailedWarmupRowsZero}{208}

\newcommand{\FixFFourFailedWarmupRowsZero}{46}

\newcommand{\FixFZeroFailedExecutionLag}{258}

\newcommand{\FixFFourFailedExecutionLag}{33}

\newcommand{\FixFZeroRan}{451}

\newcommand{\FixFFourRan}{263}

\newcommand{\FixSmallGainModel}{qwen3-4b}
\newcommand{\FixSmallGainFZero}{1}
\newcommand{\FixSmallGainFFour}{10}
\newcommand{\FixFourTwoMaxDiff}{2}

\newcommand{\FrozenRepeat}{12}
\newcommand{\FrozenCheck}{22}
\newcommand{\FrozenExplain}{37}
\newcommand{\FrozenArm}{360}

\newcommand{\SNoLibEpisodes}{90}
\newcommand{\SNoLibNoImprove}{4}

\newcommand{\SNoLibCheckFeedback}{16}

\newcommand{\EpisodeDrop}{95}%
\newcommand{\EpisodeShort}{25}%
\newcommand{\EpisodeWrong}{10}%

\newcommand{\EpisodeCaption}{%
One episode: a one-line bug that runs without error, caught by a check and confirmed by a grader the method
never sees. (a)~Amazon's 20-for-1 split on 6 June 2022 (real closes, indexed): raw quotes fall
\EpisodeDrop\% in one day. Run~1 of (b) reads the raw quotes, so its momentum signal is short Amazon
for the \EpisodeShort{} sessions after the split; on \EpisodeWrong{} of them (shaded) the task's
independent reference is long. (b)~On task \texttt{momentum\_a} the judge runs three checks.
Run~1 fails \texttt{split\_representation}; the method gets the failed check, its cause (shortened
here) and the \texttt{split\_once} rule card, never the grade, and retries within its budget of
\BudgetAttempts{} runs. Run~2 changes that one line and passes. Only then, behind the wall, does
the hidden grader compare the final program with the reference; its verdict goes to the
benchmark alone. These are the judge's real verdicts on every synthetic seed and on the real
panel. (c)~The third check on a synthetic panel: the library's lagged inverse-volatility
positions first move at row $k{+}1$ and pass; read one row early, they move at row $k$ and
fail.}

\renewcommand{\EpisodeCaption}{%
A one-line error, caught and repaired. (a)~Amazon's 20-for-1 split on 6 June 2022 reduces its
raw quote by \EpisodeDrop\%. Reading raw quotes makes the momentum program short for
\EpisodeShort{} sessions, although the reference is long on \EpisodeWrong{} (shaded).
(b)~Three default checks test \texttt{momentum\_a}. The split-equivalence failure returns
its cause (shortened here) and the split-adjustment card. The agent repairs one line on its
second of at most \BudgetAttempts{} runs. Only the benchmark receives the hidden grade.
These verdicts hold across every synthetic seed and the real panel.
(c)~On synthetic prices, lagged inverse-volatility positions first change at $k{+}1$ and
pass the timing check; positions read one row early change at $k$ and fail.}
\newcommand{\OverviewCaption}{%
What is in \lib{} and how an agent uses it. Left: the library as a cabinet whose tray holds one
piece per item, in the real counts (a book per skill, grouped by plugin; the rule cards; a cube
per check; a token per self-improvement method), with the engine on its front. Middle: an agent
is the user's own model plus the pieces it picks, which rise from the tray; it is reached through
MCP, tool schemas or the Python API. Right: on each task the agent writes a program and the
checks run; a failed check goes to the self-improvement method before the next run, within one
budget for every method, and a program that passes is delivered. Figure~\ref{fig:episode}
follows one episode.}

\renewcommand{\OverviewCaption}{%
\lib{} at a glance. Left: the library, one piece per skill, rule card, check and
self-improvement method; coloured pieces illustrate a selection. Middle: the agent is the
user's own model plus the pieces it picks or that \texttt{route()} names, used through MCP,
tool schemas or Python.
Right: a stock-strategy backtest illustrates buying and selling on historical prices and
tracking returns; the curves are schematic. The agent writes a program and the checks run;
a failure goes to the self-improvement
method before the next attempt, within one budget for every method. Only the delivered program
meets the hidden grader. The benchmark retains valid, accepted-but-wrong and rejected episodes.
Appendix Figure~\ref{fig:episode} follows one episode.}

\graphicspath{{shared/}}
\makeatletter\def\input@path{{shared/}}\makeatother %
\definecolor{codebg}{HTML}{F3F6FA}
\definecolor{codeink}{HTML}{26364A}
\definecolor{codeblue}{HTML}{245C9F}
\definecolor{codegreen}{HTML}{287C58}
\providecommand{\href}[2]{#2} %
\newcommand{\lib}{\textsc{Fin-RSI}}
\title{\lib{}: A Unified Library of Financial Skills,\\Checks, and Self-Improvement for LLM Agents}
\author{\small
\makebox[\dimexpr\textwidth-5mm\relax][s]{Wanghao~Ye\textsuperscript{1} \hfill Shwai~He\textsuperscript{1} \hfill Sihan~Chen\textsuperscript{2} \hfill Yiting~Wang\textsuperscript{1} \hfill Bowei~Tian\textsuperscript{1} \hfill Guoheng~Sun\textsuperscript{1} \hfill Ziyi~Wang\textsuperscript{3} \hfill Ziyao~Wang\textsuperscript{1}}\\
\makebox[\dimexpr\textwidth-5mm\relax][s]{Yexiao~He\textsuperscript{1} \hfill Zheyu~Shen\textsuperscript{1} \hfill Meng~Liu\textsuperscript{1} \hfill Yuning~Zhang\textsuperscript{1} \hfill Meng~Feng\textsuperscript{1} \hfill Yanhong~Qian\textsuperscript{1} \hfill Yang~Wang\textsuperscript{4}}\\
\makebox[\dimexpr\textwidth-5mm\relax][s]{Siyuan~Peng\textsuperscript{1} \hfill Yilong~Dai\textsuperscript{5} \hfill Zhenle~Duan\textsuperscript{1} \hfill Joshua~Liu\textsuperscript{2} \hfill Lang~Xiong\textsuperscript{6} \hfill Aarosh~Das\textsuperscript{1} \hfill Zhaoyi~Liu\textsuperscript{1} \hfill Daeun~Jung\textsuperscript{1}}\\
\makebox[\dimexpr\textwidth-5mm\relax][s]{Aditya~Sarkar\textsuperscript{1} \hfill Parastoo~Pilevar\textsuperscript{1} \hfill Xiaochuan~Gong\textsuperscript{1} \hfill Yancheng~Zhu\textsuperscript{1} \hfill Yingying~Fan\textsuperscript{1} \hfill Qiong~Wang\textsuperscript{1} \hfill Ang~Li\textsuperscript{1}}}
\affiliations{\small
\textsuperscript{1}University of Maryland, College Park \quad
\textsuperscript{2}Carnegie Mellon University\\
\textsuperscript{3}Texas A\&M University, College Station \quad
\textsuperscript{4}Northeastern University\\
\textsuperscript{5}The University of Alabama \quad
\textsuperscript{6}Stanford University}

\renewcommand{\bibliographystyle}[1]{}
\begin{document}
\maketitle

\begin{abstract}

As large language models (LLMs) become more capable, financial agents need tools they can
readily discover and use. Yet financial software is fragmented across libraries with different
interfaces and conventions, making it difficult to build reliable agent workflows. We present
\lib{}, a unified Python library of financial skills, checks, and self-improvement designed
for LLM agents. It brings together tools for market analysis, forecasting, portfolio
construction, and trading research through shared interfaces across models and agent
frameworks. Agents can find suitable tools from natural-language requests, reuse tested
implementations, and improve generated programs through checking and feedback. Developers
can extend the library with new tools without rebuilding integrations for each model. Across
\FixModels{} open-weight models on six trading-position tasks, tool discovery increased the
proportion of episodes producing valid programs from \FixFZeroPct\% to \FixFFourPct\%; for
models with at least 14 billion parameters, it rose from \FixLargeFZeroPct\% to
\FixLargeFFourPct\%. We release the library and benchmark as a shared resource for the
community and plan to expand its tools and task coverage as financial agents evolve.\looseness=-1

\end{abstract}

\begin{figure*}[t]
\centering
\resizebox{\textwidth}{!}{\input{figures/overview_short}}
\caption{\OverviewCaption}
\label{fig:overview}
\end{figure*}

\section{Introduction}
\label{sec:intro}

The \lib{} library (RSI for recursive self-improvement) lets language-model agents use
financial tools, checks and self-improvement methods through a common interface
(Figure~\ref{fig:overview}). Agents already analyse markets and trade
\citep{yu2024finmem,zhang2024finagent,xiao2025tradingagents,yang2024finrobot}, or propose,
implement and backtest factors and models themselves \citep{li2025rdagentq,tang2025alphaagent};
\citet{dong2025agents} survey these agents. Financial language models
\citep{wu2023bloomberggpt,yang2023fingpt} support domain analysis. Surveys
\citep{nie2024survey,chen2025ffm,liang2024foundation} cover financial and time-series
models for forecasting \citep{jin2024timellm} and portfolio decisions
\citep{hwang2025decision}.\looseness=-1

For a stock backtest, an agent needs tools that preserve financial conventions. A momentum
program reading Amazon's raw quotes across its 20-for-1 split in June 2022 sees a one-day fall
of \EpisodeDrop\%, goes short, and still produces a backtest (Appendix Figure~\ref{fig:episode}).
Applying a split twice or trading on the close just read also passes without an error.
Return forecasts from language models \citep{lopezlira2023chatgpt} face look-ahead bias
\citep{glasserman2023lookahead}, yet only about a quarter of 164 surveyed papers acknowledge
this bias \citep{kong2026bias}. Reusable implementations and checks let agents address these
conventions as part of executing a task.

\lib{} brings tool discovery, execution and correction into one workflow. Models can learn
to call tools \citep{schick2023toolformer,patil2024gorilla,qin2024toolllm}, even from documentation
\citep{hsieh2023tooldoc}. Property-based and metamorphic tests check programs without known
outputs \citep{claessen2000quickcheck,segura2016metamorphic,vikram2023pbt}. Self-improvement
methods use feedback to repair programs \citep{madaan2023selfrefine,shinn2023reflexion,chen2024selfdebug}
or learn reusable lessons \citep{zhao2024expel,suzgun2026dynamic,zhang2026ace,agrawal2026gepa};
reasoning self-correction can fail without external feedback \citep{huang2024cannot}. Code benchmarks test
programming \citep{chen2021codex,lai2023ds1000,zhuo2025bigcodebench}, while financial benchmarks
evaluate financial question answering, analysis and trading
\citep{chen2021finqa,islam2023financebench,xie2024finben,bigeard2025finance,chen2025stockbench}.
Repeated backtests can select a strategy by chance
\citep{harvey2016cross,bailey2017pbo}; our benchmark checks the program against an independent
reference. A user connects a model once through \texttt{generate(messages)} and shares that
connection across tools and methods. As PyOD unifies outlier detectors
\citep{zhao2019pyod,chen2025pyod2} for use and comparison \citep{han2022adbench}, \lib{} gives
agents reusable financial tools and compares improvement methods under one judge and budget.
We contribute:\looseness=-1
\begin{itemize}
\item \textbf{a library for agents}: unified interfaces to financial skills, algorithms and
checks, with \texttt{route()} matching a task description to a tested implementation
(\S\ref{sec:library});
\item \textbf{\NMethods{} self-improvement methods} sharing an interface, financial tasks and a resource budget
(\S\ref{sec:library});
\item \textbf{a benchmark} of \NTasks{} financial programming tasks with independent references,
a hidden grader, a held-out split by algorithm and real prices around four stock splits
(\S\ref{sec:tasks});
\item \textbf{a study} of \StudyModels{} open-weight models: with \texttt{route()}, models of
\FixSizeCut{} parameters or more were valid in \FixLargeFFourPct\% of episodes instead
of \FixLargeFZeroPct\% (\S\ref{sec:results}; Figure~\ref{fig:gains}a).
\end{itemize}

\section{Related work}
\label{sec:related}

\paragraph{Models and agents for finance.}
Financial models learn momentum and trend following
\citep{lim2019enhancing,wood2022momentum,wood2024xtrend}, rank stocks for cross-sectional
momentum \citep{poh2021ltr} and allocate portfolios end to end \citep{zhang2020deep,hwang2025signature}.
They use order books \citep{zhang2019deeplob}, stock relations
\citep{li2024master,hwang2024temporal}, deep and Transformer forecasts
\citep{lim2021survey,wen2023transformers}, and foundation models for prices
\citep{shi2026kronos} and time series \citep{liang2024foundation,jin2023largemodels,jin2024position};
off-the-shelf models do not consistently beat financial benchmarks \citep{rahimikia2025tsfm}.
Language models answer time-series questions \citep{kong2025timemqa} and explain
price patterns \citep{koa2025vta}; robustness and evaluation remain open
problems \citep{kong2025tsreasoning}. They
also supply portfolio views \citep{lee2025blm}, mine alpha factors
\citep{shi2026alphajungle,tang2025alphaagent,li2025rdagentq} and trade in teams
\citep{zhao2026contesttrade}. Generative models also mine factors \citep{shi2025alphaforge},
while reinforcement learning trains execution policies \citep{fang2021universal}.
These signals must become positions \citep{firoozye2023canonical};
\lib{} makes that conversion checkable.

\paragraph{Leakage and bias.}
Point-in-time language models avoid training on the future \citep{drinkall2024timemachine}, yet models recall pre-cutoff
economic data even when told to respect historical boundaries \citep{lopezlira2025memorization}
and hold investment preferences of their own against counter-evidence \citep{lee2025yourai}.
Some trading agents are built to respect data-availability timestamps \citep{zhao2026contesttrade}.
These works govern what a model knows; \lib{} checks what a generated program does with the data
it is given.

\paragraph{Libraries and benchmarks.}
Qlib \citep{yang2020qlib} and FinRL \citep{liu2020finrl,liu2022finrlmeta} give researchers a
quantitative workflow and trading environments. FinTSB
unifies the evaluation of financial forecasters \citep{hu2025fintsb}. \lib{} supports
the agent's workflow: finding a financial function, running it and revising a program from
explicit checks.\looseness=-1

\section{The \lib{} library}
\label{sec:library}

\paragraph{Skills.}
The \NSkills{} skill documents use the Agent Skills format \citep{agentskills} to guide agents
on financial data and tools, including changed defaults, licence changes and silent failure
modes. Each claim is dated and its verification recorded. The catalogue provides
\NAlgorithms{} tested algorithms for \NAlgorithmTasks{} task types. Given raw quotes and split
factors, \texttt{positions()} applies each split once and sets positions from the previous
close; unsupported requests raise an error naming the right call. For a rule described in
words, \texttt{route(description, generate)} asks the agent's own model to find a matching
algorithm, much as Gorilla and ToolLLM retrieve an API for a request
\citep{patil2024gorilla,qin2024toolllm} and PyOD~2 lets a model choose the detector
\citep{chen2025pyod2}. The \NCards{} rule cards pair a rule, such as \emph{apply each split once},
with its source, the checks that verify it and, where one exists, the call that follows it
(Appendix~\ref{app:library}). Rules paired with verification also guide
LLM-based hardware-code optimization \citep{wang2025symrtlo}.\looseness=-1

\begin{figure*}[t]
\centering
\includegraphics{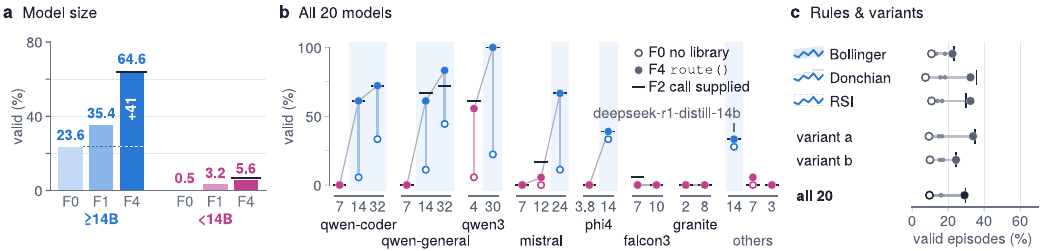}
\caption{Who gains from the library: valid episodes without it (F0), with its interface (F1),
with \texttt{route()} (F4), or with the call named (F2). (a)~Model-size groups: finding the call
helps more than listing the interface (dashed: F0; line across F4: F2). (b)~Every model, grouped
by series and size in billions (shaded: \FixSizeCut{} or more); F4 and F2 differ by at most
\FixFourTwoMaxDiff{} episodes per model. (c)~Every rule and both variants: F2 and F4 beat F0 and
F1 (grey dots: F1, F3).}
\label{fig:gains}
\end{figure*}

\paragraph{Checks.}
The \NGuards{} guards validate data, inputs and programs. Of \FixVisibleChecks{} property
checks, three run by default: finite output and two metamorphic tests
\citep{segura2016metamorphic}. Expressing the same prices as adjusted closes must preserve
the output. Changing prices from row $k$ onward must leave positions through row $k$ unchanged,
catching same-close trading (Appendix Figure~\ref{fig:episode}c). Three opt-in checks read the
task: the program must take positions, size them as specified and hold none during warm-up.
No check needs the expected output or reads the reference; every reference passes all six.
\texttt{explain()} returns causes and relevant rule cards.\looseness=-1

\paragraph{Self-improvement behind one interface.}
Every method implements \texttt{fit()} on development tasks, \texttt{solve()} on a task and
\texttt{state()}, so one can replace another without changing the agent. Inside
\texttt{solve()} a method sees the task and one judge, \texttt{check(code)}, which reports whether
the program ran and which checks failed and why, never the grade (Appendix
Figure~\ref{fig:episode}b). All methods share the same limits on model calls and program runs.
The \NMethods{} methods adapt these approaches to one interface (Appendix
Table~\ref{tab:methods}): a single attempt (\texttt{NoImprove}); retries with the error
\citep{chen2024selfdebug}, with the failed checks and their cards (\texttt{CheckFeedback}) or
with a written reflection \citep{shinn2023reflexion}; a reviewer per rule family
(\texttt{Team}); an evolved prompt \citep{agrawal2026gepa}; and a playbook learned from
development traces \citep{zhao2024expel,zhang2026ace}, kept only if no development task regresses.\looseness=-1

\paragraph{Agent interfaces.}
\lib{} installs as the Python package \texttt{fin\_skills}, whose base install needs only numpy,
pandas and scipy. One registry serves \NTools{} tools as an MCP server \citep{mcp2026spec}, as
OpenAI and Anthropic tool schemas and as Python calls, so the same agent runs on any of them,
whether it acts through tool calls \citep{yao2023react} or through code \citep{wang2024codeact}.

For the stock backtest in Figure~\ref{fig:overview}, an agent describes its strategy in words,
finds a matching algorithm with \texttt{route()} and calls \texttt{positions()} to compute
holdings from prices and split factors. For a custom strategy, \texttt{signal\_positions()}
applies the same timing and split conventions to the agent's own signal. Checks on a generated
program identify failed properties and return the relevant rule cards. The agent can then revise the program
with any of the \NMethods{} methods, using the same task, checks and budget (Appendix
\ref{app:library}). Users can change their model or improvement method while retaining the
same financial tools and workflow.

\section{Benchmark}
\label{sec:tasks}

The suite has \NTasks{} tasks: \NTaskAlgorithms{} algorithms in \NFamilies{} families, in two
variants each (Appendix Table~\ref{tab:tasks}). Each states a rule in prose and asks for
\texttt{solve(prices, split\_factors)}, which returns positions or the value known at the
previous close. Rules such as time-series momentum and minimum-variance allocation
also serve as baselines for learned strategies
\citep{lim2019enhancing,wood2022slow,hwang2025signature}; the others are common technical, risk
and forecast rules. An error in any of them carries into every strategy built on it. Every task can
be solved by hand or with one library call. Synthetic panels hold four assets over \NRows{}
business days with four splits of different ratios. Variant~a gives cumulative split factors
and sorted columns; variant~b gives per-event ratios, shuffled columns and other parameters. A
real panel with the as-traded closes of four stocks that all split within \NRealRows{} sessions
in 2022 checks every task on real prices (Appendix~\ref{app:bench}); the study uses the
synthetic panels.

A program is \emph{valid} only if it runs, passes every visible check and matches an independent
numpy reference to within $10^{-8}$ ($10^{-5}$ for minimum variance). A hidden grader makes this comparison after the program is
delivered, and its verdict never enters a model's prompt. A program that passes every check but still differs from the reference
is reported separately as \emph{accepted but wrong}. Before any model ran, we
confirmed that the references and the library agree, that correct programs pass every check, and
that a planted version of each failure fails both its check and the grade
(Appendix~\ref{app:bench}). Tasks are split by algorithm, so a held-out task asks for a rule that
no development task uses: \NDevTasks{} development and \NHeldTasks{} held-out tasks, covering all
\NFamilies{} families.

\section{Experiments}
\label{sec:study}
\subsection{Setup}

\begin{figure*}[t]
\centering
\includegraphics{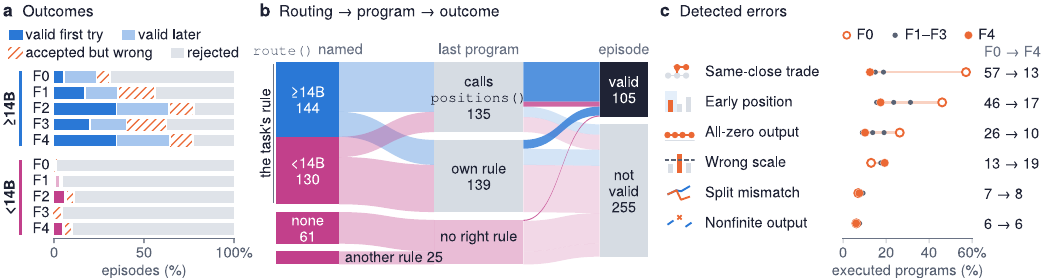}
\caption{How the library helps. (a)~How episodes ended, by arm and model size;
\emph{accepted but wrong} programs passed every check and were still wrong. (b)~The \FixArm{} F4
episodes: what \texttt{route()} named, whether the final program called \texttt{positions()},
and the outcome (band width: episodes). (c)~Share of executed programs that failed each check
(F0 open, F4 filled, F1--F3 grey). Small glyphs illustrate the failure types; the plotted
rates come from the recorded programs.}
\label{fig:analysis}
\end{figure*}

We ran \StudyModels{} open-weight instruction models from \StudyFamilies{} families, 2B to 32B
total parameters, at temperature 0.1 on six position tasks: the Bollinger, RSI and Donchian rules, two variants each, on \StudySeeds{} synthetic seeds. An
\emph{episode} is one model's attempt at one task and seed, with at most \FixAttempts{} programs;
refusals, truncated replies and programs that fail to run all count as failures. Variants and
seeds share three rules, so we treat models as the independent units.\looseness=-1

\paragraph{The library.}
Five arms differ only in what the model is told about the library and whether it may import
it. All share the prompt, the judge and the budget, with \FixEpisodesPerArm{} episodes per model
and arm. F0 has no library. F1 lists the interface of \texttt{positions()} in the system prompt.
F2 adds one sentence naming the call that solves the task, providing a reference for successful
tool discovery. F3 is F1 with each algorithm's rule stated in one line, as tool documentation
would give it \citep{hsieh2023tooldoc}. F4 asks \texttt{route()}, then uses F2 with the named
call, or F1 when no call is named.

\paragraph{Feedback and self-improvement.}
Without the library, we compared three feedback conditions on the same tasks, seeds and
models: \texttt{Repeat} retries without diagnostic feedback, \texttt{Check} names failed
checks and \texttt{Explain} adds their causes. We compared \NMethods{} improvement methods
under equal budgets on \StudyCoreModels{} core models.\looseness=-1

\subsection{Results}
\label{sec:results}

\paragraph{Finding the call drives the gain.}
Without the library (F0), \FixFZeroPct\% of the \FixArm{} episodes were valid. Listing the
interface (F1) raised this to \FixFOnePct\%, better than F0 for \FixFOneMinusFZeroBetter{}
models and worse for \FixFOneMinusFZeroWorse{} (two-sided sign test over models, $p=0.04$), but
most F1 programs used the library only to wrap a rule of their own, and
\FixFOneUsesPositions{} of \FixFOnePrograms{} called \texttt{positions()}. Stating
every rule (F3) did not change this (\FixFThreePct\%). Naming the call (F2) raised validity to
\FixFTwoPct\%, better than F0 for \FixFTwoMinusFZeroBetter{} models and worse for none
($p<0.001$), and better than F1 for \FixFTwoMinusFOneBetter{} and worse for
\FixFTwoMinusFOneWorse{} ($p=0.27$). With \texttt{route()} the models found the call themselves:
it named the task's rule in \FixRouteRight{} of the \FixArm{} F4 episodes
(Figure~\ref{fig:analysis}b), and F4 was valid in \FixFFourPct\% of episodes, within one point of
F2, beating F0 for \FixFFourMinusFZeroBetter{} models without a loss. Each rule and both
variants improve (Figure~\ref{fig:gains}c).\looseness=-1

\paragraph{Larger models gain the most.}
The \FixLargeModels{} models of \FixSizeCut{} parameters or more routed every episode to the
right rule and went from \FixLargeFZeroPct\% to \FixLargeFFourPct\% valid, level with F2
(\FixLargeFTwoPct\%). All \FixLargeModels{} improved, six by seven episodes or more. Within each series
with both size groups, models of \FixSizeCut{} or more gained more than the smaller models
(Figure~\ref{fig:gains}b). Among \FixSmallModels{} smaller models, \texttt{route()} correctly
routed \FixSmallFFourTold{} episodes; \texttt{\FixSmallGainModel} rose from
\FixSmallGainFZero{} to \FixSmallGainFFour{} valid episodes out of \FixEpisodesPerArm{}.\looseness=-1

\paragraph{The library reduces timing errors, and feedback repairs programs.}
Without the library, \FixFZeroFailedExecutionLag{} of the \FixFZeroRan{} programs that ran
traded on the close they read and \FixFZeroFailedWarmupRowsZero{} took positions during
warm-up; with \texttt{route()}, \FixFFourFailedExecutionLag{} and \FixFFourFailedWarmupRowsZero{}
of \FixFFourRan{} did (Figure~\ref{fig:analysis}c), and the checks rejected every one of them.
Retries recovered many episodes: in F4,
\the\numexpr\FixFFourValid-\FixFFourFirst\relax{} of the \FixFFourValid{} valid episodes came
after a failed first program (Figure~\ref{fig:analysis}a).

\paragraph{Richer feedback recovers more episodes.}
Without the library, validity increased with richer feedback:
\FrozenRepeat{}, \FrozenCheck{} and \FrozenExplain{} of \FrozenArm{} with \texttt{Repeat},
\texttt{Check} and \texttt{Explain} (Appendix Table~\ref{tab:frozen}). With failed checks and
rule cards, \texttt{CheckFeedback} and \texttt{Evolve} quadrupled single-attempt
validity on the core models (\SNoLibCheckFeedback{} against \SNoLibNoImprove{} of \SNoLibEpisodes{} episodes).\looseness=-1

These results point to a practical requirement for agent libraries: the model must be able to
find the implementation that fits its task. The interface alone gave a smaller gain, even
when it described every rule. With \texttt{route()}, agents selected the call themselves and
approached the baseline in which it was supplied. Pairing that discovery step with executable
checks gives agents both a tested implementation to use and specific feedback for revising
their own programs.

\section{Conclusion}
\label{sec:conclusion}

\lib{} is a library for agents: one set of interfaces connects different language models to
financial tools, checks and self-improvement methods. Across \StudyModels{} models, finding
the matching call raised pooled validity beyond listing the interface. As agents become more
capable, this common interface lets them reuse tested implementations and act on feedback,
supporting~reliable~financial~programs.

\section{Broader impact}
\label{sec:impact}

Unified interfaces make a new tool or check reusable across models, reducing the need for
separate integrations. Researchers can extend the same library as models and tasks evolve.
The shared benchmark, equal budgets and public run records support reproducible comparisons
and the study of failures. Together, they give researchers a common base for building tools
for increasingly capable financial agents.

\paragraph{Limitations.}
\label{sec:limits}
Our study covers position rules and open-weight models up to 32B. The same interface supports
other task families and larger models.\looseness=-1

\appendix
\raggedbottom

\section{The library in use}
\label{app:library}

\begin{figure*}[!t]
\centering
\input{shared/figures/episode}
\caption{\EpisodeCaption}
\label{fig:episode}
\end{figure*}

Figure~\ref{fig:episode} follows a stock-backtest program through the workflow in
Figure~\ref{fig:overview}. The agent's momentum program runs without an error but mistakes a
stock split for a price collapse. A failed check points to the relevant rule; a one-line
revision then passes. The agent receives an explanation of the failure while the independent
grade stays hidden.\looseness=-1

The same model connection supports each improvement method. This example uses Reflexion
with the library on a benchmark task; \texttt{generate} wraps the user's model:
\begin{lstlisting}[language=Python,basicstyle=\ttfamily\small\color{codeink}]
from fin_skills import rsi
from fin_skills.tasks import suite

def generate(messages):
    return my_model.chat(messages)

tasks = suite()
method = rsi.Reflexion(library=True)
method.fit(generate, tasks.dev)
episode = method.solve(
    generate, tasks.heldout[0], seed=11)
\end{lstlisting}
An agent finds an algorithm with \texttt{route()} and passes prices and split factors to
\texttt{positions()}, which handles execution timing, warm-up and scale.
\texttt{signal\_positions()} applies the same conventions to an agent's own rule;
\texttt{rolling()} provides volatility, risk and forecasts using information available at
the previous close.\looseness=-1

The \NCards{} rule cards explain execution timing, causal features, split adjustment and
conventions, warm-up and the first return. Each task receives the applicable cards;
\texttt{explain()} retrieves the ones that address a failed check. This gives the agent
guidance tied to the program it has just written.

\section{Learning from failed programs}
\label{app:methods}

\begin{table}[!ht]
\centering\small
\setlength{\tabcolsep}{3pt}
\begin{tabular}{@{}>{\raggedright\arraybackslash}p{2.5cm}>{\raggedright\arraybackslash}p{5.3cm}@{}}
\toprule
Method & How it improves the program \\
\midrule
NoImprove & One program, without feedback. \\
SelfDebug & Retry on execution errors\newline until the program runs. \\
CheckFeedback & Retry with failed checks,\newline causes and their rule cards. \\
Reflexion & Add a written reflection before each retry. \\
Team & Use a writer and one reviewer\newline assigned to each rule family. \\
PlaybookLearning & Add learned entries on tool use, retrieval and experience to check feedback. \\
Evolve & Evolve a prompt over \EvolveGenerations{} generations\newline using \EvolveMinibatch{} development tasks; then\newline add it to the failed-check feedback. \\
\bottomrule
\end{tabular}
\caption{Seven ways to improve a program under the same prompt, judge and budget. These are
adaptations of the cited approaches, using one library interface.}
\label{tab:methods}
\end{table}

Table~\ref{tab:methods} brings debugging \citep{chen2024selfdebug}, reflection
\citep{shinn2023reflexion}, multi-role review, playbooks \citep{zhao2024expel,zhang2026ace}
and prompt evolution \citep{agrawal2026gepa} into one interface. Team uses the same model to
write and review programs, with each reviewer assigned a rule family. PlaybookLearning keeps
lessons that preserve performance on every development task; Evolve keeps a prompt addition
when it improves the development score. The learned state stays fixed during testing.

For comparison, Repeat, Check and Explain vary only the feedback after a failed attempt.
They share the prompt and budget, and deliver the first program to pass the visible checks.
Table~\ref{tab:frozen} reports the no-library results.

\begin{table}[!ht]
\centering\small\setlength{\tabcolsep}{2pt}
\begin{tabular}{@{}lrrrrrr@{}}
\toprule
Method & \shortstack{qwen-\\coder-\\32b} & \shortstack{mistral-\\small-\\24b} & \shortstack{granite-\\8b} & \shortstack{olmo3-\\7b} & \shortstack{phi4-\\mini} & all \\
\midrule
NoImprove & 1 & 3 & 0 & 0 & 0 & 4 \\
SelfDebug & 1 & 3 & 0 & 0 & 0 & 4 \\
CheckFeedback & 7 & 9 & 0 & 0 & 0 & 16 \\
Reflexion & 5 & 6 & 0 & 0 & 0 & 11 \\
Team & 2 & 5 & 0 & 0 & 0 & 7 \\
Evolve & 7 & 9 & 0 & 0 & 0 & 16 \\
PlaybookLearning & 7 & 8 & 0 & 0 & 0 & 15 \\
\bottomrule
\end{tabular}

\caption{Valid episodes without the library, out of \FixEpisodesPerArm{} per core model.
Across all \StudyModels{} models and \FrozenArm{} episodes per arm, Repeat was valid in
\FrozenRepeat, Check in \FrozenCheck{} and Explain in \FrozenExplain.}
\label{tab:frozen}
\end{table}

\section{Tasks, prices and evaluation}
\label{app:bench}

The task suite covers \NTaskAlgorithms{} algorithms in five families, with two variants of
each rule (Table~\ref{tab:tasks}). Development includes \NForcedDev{} fixed algorithms and
\NDrawnDev{} of the remaining \NDrawn{}, selected with a seed. Synthetic panels use seeded
Gaussian returns and four splits at fixed rows: 2-for-1, 3-for-1, 3-for-2 and 1-for-2.
The real panel spans \NRealRows{} sessions, \RealFirst{} to \RealLast{}. It reconstructs
as-traded quotes from adjusted closes and a split table checked against company announcements;
the reconstruction code is released.

\begin{table}[!ht]
\centering\small
\setlength{\tabcolsep}{3pt}
\begin{tabular}{@{}>{\raggedright\arraybackslash}p{1.2cm}>{\raggedright\arraybackslash}p{3.0cm}>{\raggedright\arraybackslash}p{3.5cm}@{}}
\toprule
Family & Development & Held out \\
\midrule
Portfolio & Cross-sectional momentum; inverse volatility & Hierarchical risk parity; minimum variance \\
Signal & Bollinger reversion; moving-average crossover; MACD; momentum; volatility-targeted momentum & Donchian breakout; RSI reversion \\
Volatility & Historical volatility & Exponentially weighted volatility \\
Risk & -- & Historical VaR/ES; normal VaR/ES \\
Forecast & Mean & Drift \\
\bottomrule
\end{tabular}
\caption{The 17 algorithms and their development split. Each has two task variants.
Portfolio and signal tasks use \texttt{positions()}; the other families use \texttt{rolling()}.}
\label{tab:tasks}
\end{table}

Visible checks provide feedback, while an independent reference grades each program with
tolerance $10^{-8}$ ($10^{-5}$ for minimum variance). Before evaluating models, we confirmed
that the reference and library agree on every task and panel, within $\RefGapOther$
($\RefGapMinVariance$ for minimum variance). Correct programs passed every check; planted
errors failed both their checks and the independent grade. Programs run in separate
processes with time and memory limits.\looseness=-1

\section{How models used the library}
\label{app:pfix}

\begin{figure}[!ht]
\centering
\includegraphics[trim=0 1bp 0 4bp,clip]{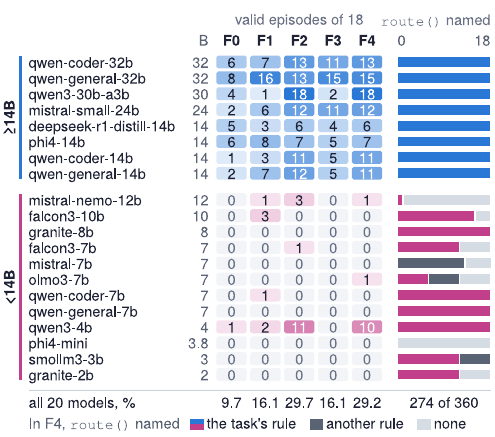}
\caption{Valid episodes out of \FixEpisodesPerArm{} for every model and arm. Darker cells
indicate more valid episodes; blue marks models of \FixSizeCut{} or more, magenta the smaller
models. The bars show what \texttt{route()} selected in F4. The bottom row gives the overall
percentages and correct selections out of \FixArm{}. B denotes model size~in~billions~of~parameters.}
\label{fig:matrix}
\end{figure}

Figure~\ref{fig:matrix} retains all \FixModels{} models under every condition. F1 lists the
\FixInterfaceAlgorithms{} algorithms accepted by \texttt{positions()} and permits importing
\texttt{fin\_skills}. F2 also names the call that implements the task's rule. In F4,
\texttt{route()} selects that name and counts toward the episode's budget. Each arm allows
\FixAttempts{} programs, \FixCalls{} model calls and \FixTokens{} generated tokens per
episode, with a limit of \FixOutputCap{} per reply. All six property checks provide Check feedback.\looseness=-1

Choosing a tested implementation mattered more than importing the package. Programs
imported \texttt{fin\_skills} in \FixFOneUsesImportsFinSkills{} of \FixFOnePrograms{} cases
in F1 and \FixFFourUsesImportsFinSkills{} of \FixFFourPrograms{} in F4. When the final F4
program called \texttt{positions()}, \FixFFourLastUsesPositionsValid{} of
\FixFFourLastUsesPositions{} episodes were valid. When it passed its own rule to
\texttt{signal\_positions()}, \FixFFourLastUsesSignalPositionsValid{} of
\FixFFourLastUsesSignalPositions{} were valid.

\newpage

The separate feedback study used the three default checks and \StudyOutputCap{} output
tokens per call. It compared feedback across \StudyModels{} models and the \NMethods{}
methods on \StudyCoreModels{} core models, with applicable cards in the first prompt
(Table~\ref{tab:frozen}). Each program ran in isolation, with network and subprocess access disabled.

\end{document}